\documentclass[lettersize,journal]{IEEEtran}
\usepackage{amsmath,amsfonts}
\usepackage{algorithmic}
\usepackage{algorithm}
\usepackage{array}
\usepackage{graphicx}
\usepackage{xcolor}
\usepackage{cite}
\usepackage{url}
\usepackage{hyperref}
\usepackage{booktabs}
\usepackage{multirow}
\usepackage{caption}
\usepackage{subcaption}
\usepackage{bm}
\usepackage{amsthm}
\usepackage{color}
\usepackage{mathtools}
\usepackage{times}
\usepackage{booktabs}
\usepackage{multirow}   

\hypersetup{
 colorlinks=true,
 citecolor=blue,
 linkcolor=red,
 urlcolor=blue,
 pdfpagemode=UseNone,
 pdfstartview=FitH}
 
\usepackage[all]{hypcap}

\begin{document}

\title{Line Space Clustering (LSC): Feature-Based Clustering using K-medians and Dynamic Time Warping for Versatility}

\author{
    Joanikij Chulev\\
    \href{mailto:joanikij.chulev@proton.me}{joanikijchulev@proton.me}\\
    Universitat Politècnica de Catalunya · Barcelona Tech - UPC\\[1ex]
    Angela Mladenovska\\
    \href{mailto:amladenovska001@gmail.com}{amladenovska001@gmail.com}\\
    Ss. Cyril and Methodius University · UKIM
}

\markboth{LSC: Line Space Clustering, January~2025}%
{Chulev \MakeLowercase{\textit{et al.}}: Line Space Clustering (LSC)}

\maketitle

\begin{abstract}
Clustering high-dimensional data is a critical challenge in machine learning due to the curse of dimensionality and the presence of noise. Traditional clustering algorithms often fail to capture the intrinsic structures in such data. This paper explores a combination of clustering methods, which we called Line Space Clustering (LSC), a representation that transforms data points into lines in a newly defined feature space, enabling clustering based on the similarity of feature value patterns, essentially treating features as sequences.  LSC employs a combined distance metric that uses Euclidean and Dynamic Time Warping (DTW) distances, weighted by a parameter $\alpha$, allowing flexibility in emphasizing shape or magnitude similarities. We delve deeply into the mechanics of DTW and the Savitzky-Golay filter, explaining their roles in the algorithm. Extensive experiments demonstrate the efficacy of LSC on synthetic and real-world datasets, showing that randomly experimenting with time-series optimized methods sometimes might surprisingly work on a complex dataset, particularly in noisy environments. Source code and experiments are available at: \href{https://github.com/JoanikijChulev/LSC}{GitHub}.

\end{abstract}


\section{Introduction}
\IEEEPARstart{C}{lustering} is a fundamental and widely studied task within unsupervised machine learning, aiming to group similar data points based on inherent patterns without requiring labeled examples \cite{Xu2015SurveyClustering}. This approach is particularly beneficial in some real-world scenarios when labeled data is either scarce or expensive to obtain. Clustering algorithms have been successfully applied in various domains, including image segmentation, where grouping pixels based on similarities improves object recognition and visual interpretation \cite{Zhang2021ImageSegmentationReview}.

Despite broad applications, clustering high-dimensional data remains a significant challenge due to the well-known \emph{curse of dimensionality}. As dimensionality increases, data points tend to distribute sparsely across feature spaces, causing distances between points to become less meaningful and increasingly uniform, thus weakening the discriminative power of traditional distance metrics \cite{Steinbach2004HighDimensional}. High-dimensional datasets often contain substantial noise, redundancy, and intricate feature interactions, making it more challenging to identify true underlying cluster structures \cite{Zimek2012SurveyHighDimClustering}. Noise further exacerbates this problem by masking relevant patterns and amplifying irrelevant features, reducing the accuracy and reliability of clustering results.

Traditional clustering algorithms such as K-means \cite{Arthur2007KMeansPlusPlus} and hierarchical clustering \cite{Murtagh2012HierarchicalReview} rely predominantly on straightforward distance calculations, which often inadequately capture the complex interactions and non-linear patterns characteristic of high-dimensional spaces. These algorithms may fail to uncover meaningful groupings in datasets where proximity in Euclidean space does not accurately reflect genuine similarity among data points. Consequently, there is a compelling need to experiment with clustering methodologies capable of effectively managing dimensionality-related issues, better handling noise, and capturing the relationships, if any, present in high-dimensional data environments.

\subsection{Motivation}
The foundational premise of Line Space Clustering (LSC) involves interpreting each data instance not as a static entity within conventional high-dimensional feature spaces, but as an ordered sequence of feature values that, when visualized against corresponding feature indices, manifest as distinctive trajectories or curves. This sequential conceptualization transforms the analytical focus from isolated dimensional attributes to the recognition and examination of holistic patterns and relationships embedded within the data. Consequently, LSC is uniquely positioned to discern nuanced interactions among features, revealing structures typically obscured by standard vector-centric methodologies.
In numerous applied contexts, scrutinizing the collective pattern exhibited by feature values across multiple dimensions yields deeper insight than evaluations based solely on isolated attribute magnitudes. By embodying data points as linearly sequenced entities, LSC facilitates the application of sophisticated sequence alignment methods, notably Dynamic Time Warping (DTW), thereby enhancing the capability to perform comparisons and pattern-driven clustering based on the chance that some unseen temporal dynamics may be inherent in some complex datasets.

\section{Related Work}
Clustering high-dimensional datasets has been the subject of extensive research, with diverse methodologies developed to address the intrinsic complexities introduced by dimensionality and noise.

\subsection{Distance-Based Clustering}
Traditional clustering algorithms, such as K-means \cite{Arthur2007KMeansPlusPlus} and hierarchical clustering \cite{Murtagh2012HierarchicalReview}, typically utilize distance measures like the Euclidean metric to determine similarities among data points. Nevertheless, the utility of such distance metrics diminishes in high-dimensional spaces due to the phenomenon of distance concentration, wherein distances between data points become increasingly uniform, reducing discriminative capability \cite{Steinbach2004HighDimensional}. Studies highlight the diminished effectiveness of distance metrics in distinguishing clusters as dimensionality escalates \cite{Zimek2012SurveyHighDimClustering}.

\subsection{Subspace and Projected Clustering}
To alleviate the curse of dimensionality, subspace clustering \cite{Parsons2004SubspaceReview} and projected clustering methods \cite{Aggarwal1999ProjectedClustering} identify meaningful clusters confined within specific subsets of dimensions, thus mitigating the detrimental effects of irrelevant or noisy features. Prominent examples include CLIQUE \cite{Agrawal2005CLIQUE} and PROCLUS \cite{Aggarwal1999ProjectedClustering}. Despite their efficacy in identifying clusters in sub-dimensional spaces, these approaches typically involve computationally intensive processes for optimal subspace selection and can be vulnerable to performance degradation in highly noisy contexts \cite{Parsons2004SubspaceReview}.

\subsection{Spectral Clustering}
Spectral clustering methods leverage eigenvector decomposition of similarity matrices to perform dimensionality reduction prior to clustering. They construct a graph representation of data points and identify clusters via spectral decomposition of the graph Laplacian \cite{VonLuxburg2007SpectralClustering}. Spectral clustering is highly effective in capturing complex structures; however, its performance significantly depends on the choice of similarity metrics and parameters used for scaling. Additionally, constructing similarity matrices can incur substantial computational overhead for large-scale datasets \cite{Shi2000SpectralClustering}.

\subsection{Density-Based Clustering}
Density-based clustering algorithms, notably DBSCAN \cite{Ester1996DBSCAN}, define clusters based on dense regions within the data space, enabling the detection of clusters of arbitrary shapes and providing inherent robustness to noise. Extensions like OPTICS \cite{Ankerst1999OPTICS} further enhance flexibility by ranking points to handle clusters with varying densities. Nevertheless, these methods necessitate careful parameter tuning, and their effectiveness diminishes significantly when confronted with high-dimensional data, due to the increasing difficulty of reliable density estimation \cite{Zimek2012SurveyHighDimClustering}.

\subsection{Time-Series Clustering with DTW}
Dynamic Time Warping (DTW) is a prominent method utilized in time-series analysis, providing an elastic measure to align sequences non-linearly, thereby accounting for temporal variations \cite{Aghabozorgi2015TimeSeriesClustering}. DTW has demonstrated robust performance in time-series clustering despite temporal distortions \cite{Keogh2005DTWReview}. However, the broader applicability and papers of DTW to static high-dimensional data has been limited, underscoring the need for methodological adaptations or novel integrations to effectively extend its benefits.

\section{Methodology}

\subsection{Line Space Transformation}
In traditional clustering, data points are considered as vectors in a $d$-dimensional space, where $d$ is the number of features. However the idea is that, this perspective may not capture the underlying patterns within the feature values of each data point, even in static data.

To overcome this limitation, we introduce the concept of \emph{line space}, where each data point is represented as a line or sequence of feature values plotted against the feature indices. This transformation allows us to analyze the pattern and shape of the feature values within individual data points.

Given a dataset $\mathbf{X} \in \mathbb{R}^{n \times d}$ with $n$ samples and $d$ features, we represent each data point $\mathbf{x}_i$ as a line in the line space:

\begin{equation}
\mathbf{L}_i = \{(f_j, x_{ij}) \mid j = 1, 2, \dots, d\}
\end{equation}

where $f_j$ is the feature index (treated as a sequential index similar to time in time-series data), and $x_{ij}$ is the value of feature $j$ for data point $i$.

This representation effectively converts the static high-dimensional data point into a sequence, enabling the application of sequence alignment and comparison techniques.
The line space representation offers several advantages:

\begin{itemize}
    \item \textbf{Pattern Recognition}: By treating feature indices as a sequence, we can recognize patterns and trends within individual data points.
    \item \textbf{Alignment Flexibility}: Using sequence alignment techniques like DTW, we can compare data points even if they have non-linear shifts or distortions in feature values.
    \item \textbf{Noise Reduction}: Smoothing techniques can be applied to the lines to reduce noise while preserving essential patterns.
    \item \textbf{Dimensionality Reduction}: Analyzing patterns may capture the data's intrinsic structure better than considering all features independently.
\end{itemize}

\begin{figure}[h]
\centering
\includegraphics[width=3.8 in]{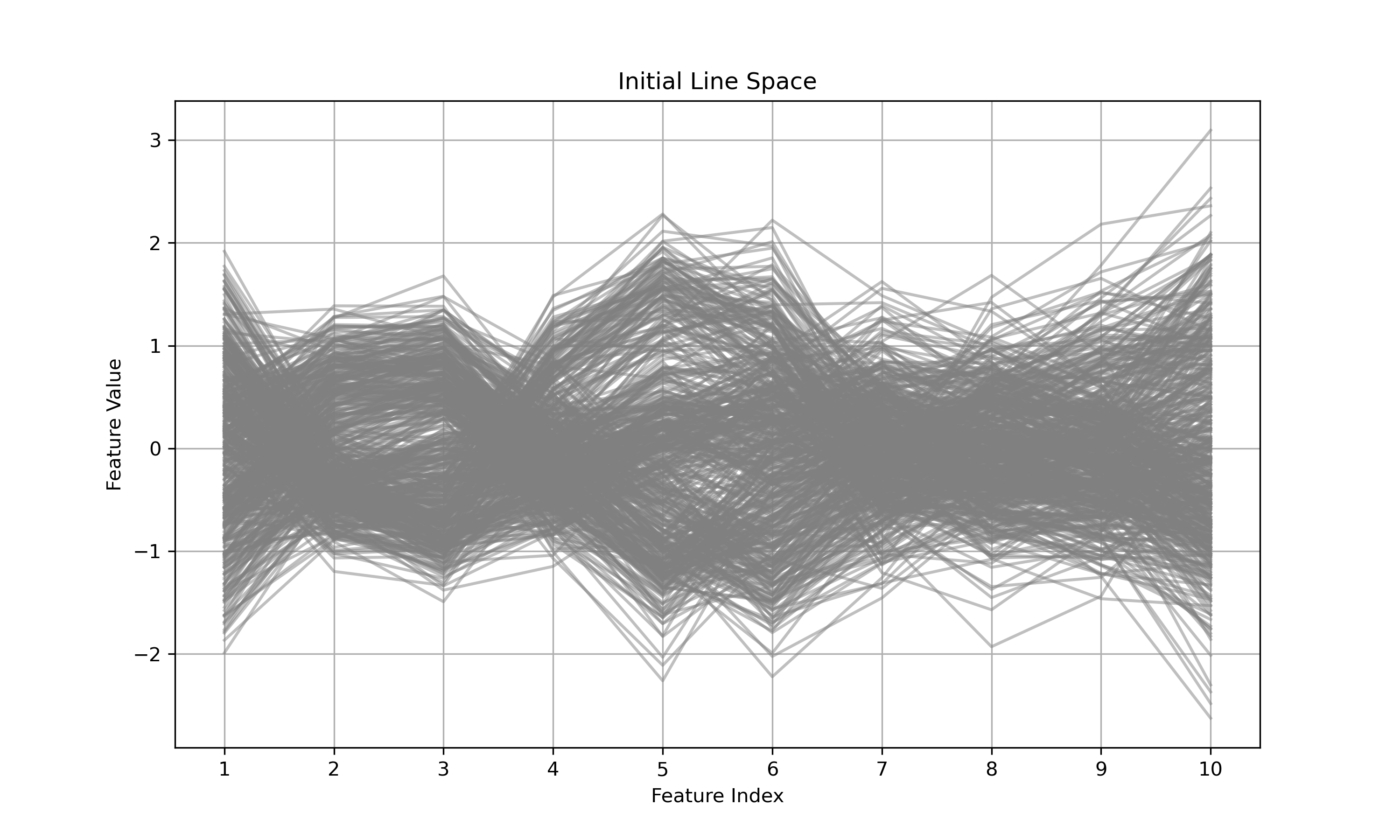}
\caption{Visualization of data points in the line space. Each line represents a data point plotted against feature indices.}
\label{fig:line_space}
\end{figure}

\subsection{Dynamic Time Warping (DTW)}
\subsubsection{Overview}
Dynamic Time Warping (DTW) is an established algorithm extensively utilized for quantifying the similarity between sequences, particularly when they exhibit temporal variations such as stretching, compression, or phase shifts \cite{Keogh2005DTWReview}. Initially developed for applications in speech recognition, DTW dynamically aligns sequences to minimize an overall distance metric, thus effectively handling misalignments in sequence indexing.

\subsubsection{Application of DTW in Line Space}
In the proposed Line Space representation, DTW serves to quantify similarities between data points by aligning their sequences of feature values. This capability is crucial because it allows the identification of underlying structural similarities even when the corresponding sequences are not aligned linearly or exhibit non-linear distortions. DTW's flexibility in sequence alignment makes it particularly effective for applications where the general pattern or shape of feature variations across dimensions is more informative than absolute feature magnitudes, especially in high-dimensional and noisy scenarios \cite{Aghabozorgi2015TimeSeriesClustering}.

For instance, two data points may possess similar underlying feature trends but differ slightly in the values (distances) at which these trends occur. Traditional Euclidean distance metrics might incorrectly interpret these as significantly dissimilar, whereas DTW identifies and aligns these patterns to accurately capture their true similarity \cite{Berndt1994DTWOriginal}.

\subsubsection{Computing DTW Distance}
The DTW distance between two sequences $\mathbf{s} = (s_1, s_2, \dots, s_T)$ and $\mathbf{t} = (t_1, t_2, \dots, t_{T'})$ is computed by constructing a cost matrix $D$ where each element $D(i,j)$ represents the cumulative cost of aligning $s_1$ to $s_i$ with $t_1$ to $t_j$.

The recursive formula for $D(i,j)$ is:

\begin{equation}
D(i,j) = d(s_i, t_j) + \min \begin{cases}
D(i-1,j), \\
D(i,j-1), \\
D(i-1,j-1)
\end{cases}
\end{equation}

where $d(s_i, t_j)$ is the local cost (distance) between elements $s_i$ and $t_j$, typically computed as:

\begin{equation}
d(s_i, t_j) = |s_i - t_j|
\end{equation}

The boundary conditions are set as:

\begin{align}
D(0,0) &= 0 \\
D(i,0) &= D(i-1,0) + d(s_i, t_0) \quad \text{for } i > 0 \\
D(0,j) &= D(0,j-1) + d(s_0, t_j) \quad \text{for } j > 0
\end{align}

The DTW distance is then the cumulative cost at $D(T,T')$, representing the optimal alignment cost between the two sequences.

\subsubsection{Warping Path}
To describe global alignment between sequences X and Y, consider a set of index pairs that meet specific criteria. This leads to the concept of a warping path. A warping path $\mathcal{W}$ is a sequence of matrix elements that defines a mapping between the sequences:

\begin{equation}
\mathcal{W} = \{ (i_1, j_1), (i_2, j_2), \dots, (i_K, j_K) \}
\end{equation}

subject to boundary conditions, continuity, and monotonicity constraints:

\begin{itemize}
    \item \textbf{Boundary Conditions}: $i_1 = 1$, $j_1 = 1$, $i_K = T$, $j_K = T'$
    \item \textbf{Continuity}: $i_{k+1} - i_k \leq 1$, $j_{k+1} - j_k \leq 1$
    \item \textbf{Monotonicity}: $i_{k+1} - i_k \geq 0$, $j_{k+1} - j_k \geq 0$
\end{itemize}

The optimal warping path minimizes the total cost:

\begin{equation}
DTW(\mathbf{s}, \mathbf{t}) = \min_{\mathcal{W}} \sum_{k=1}^{K} d(s_{i_k}, t_{j_k})
\end{equation}

\subsubsection{Time Complexity}
The classical DTW algorithm inherently possesses a quadratic time complexity, specifically $O(T \times T')$, where $T$ and $T'$ represent the lengths of the two sequences being compared. This computational demand renders DTW impractical for large datasets or lengthy sequences. To address this computational limitation, we employ the FastDTW algorithm \cite{Salvador2007FastDTW}, an approximation method designed to significantly reduce computational overhead. FastDTW achieves a linear approximation of DTW by progressively refining sequence alignment at coarser resolutions before incrementally enhancing alignment precision at finer scales, thus substantially enhancing scalability and efficiency for practical applications.

\subsection{Savitzky-Golay Filter}
\subsubsection{Overview}
The Savitzky-Golay filter \cite{Savitzky1964Smoothing} is a digital signal processing technique designed to smooth data by applying polynomial regression through successive subsets of adjacent data points using a linear least squares methodology. Distinct from simpler smoothing techniques such as moving averages, the Savitzky-Golay filter excels at preserving essential signal characteristics, including peak amplitude, width, and the overall shape profile. This capability renders it particularly suitable for applications involving noisy datasets where maintaining the integrity and defining features of the original signal is critical \cite{Schafer2011SavitzkyGolayReview}.

\subsubsection{Mathematical Formulation}
The filter operates by moving a window of size $2m + 1$ (where $m$ is the window half-width) across the data and fitting a polynomial of degree $n$ to the data points within the window. The central point is then replaced with the value of the polynomial at that point.

Let $\mathbf{y} = [y_{i-m}, \dots, y_i, \dots, y_{i+m}]^T$ be the vector of data points within the window centered at $i$. The goal is to find polynomial coefficients $\mathbf{a} = [a_0, a_1, \dots, a_n]^T$ that minimize the squared error:

\begin{equation}
\min_{\mathbf{a}} \sum_{k=-m}^{m} \left( y_{i+k} - \sum_{j=0}^{n} a_j (k)^j \right)^2
\end{equation}

This is a linear least squares problem that can be solved efficiently.

\subsubsection{Filter Coefficients}
The filter coefficients are calculated once for a given window size and polynomial degree and applied to all data points. The coefficients depend only on the relative positions within the window and can be precomputed.

\subsubsection{Parameter Selection}
The choice of window size and polynomial degree affects the filter's performance:

\begin{itemize}
    \item \textbf{Window Size}: Larger windows provide more smoothing but may distort features.
    \item \textbf{Polynomial Degree}: Higher-degree polynomials can fit more complex shapes but may overfit noise.
\end{itemize}

In our experiments, we set the window length to 5 and the polynomial order to 2, trying to balancing smoothing and feature preservation. Although, more testing would be promising.

\subsubsection{Application in LSC}
In LSC, the Savitzky-Golay filter is applied to each data line to reduce noise before computing distances. This enhances the the algorithm, especially in noisy environments, by reducing distances between point outliers in the line space.

\begin{figure}[h]
\centering
\includegraphics[width=3.4in]{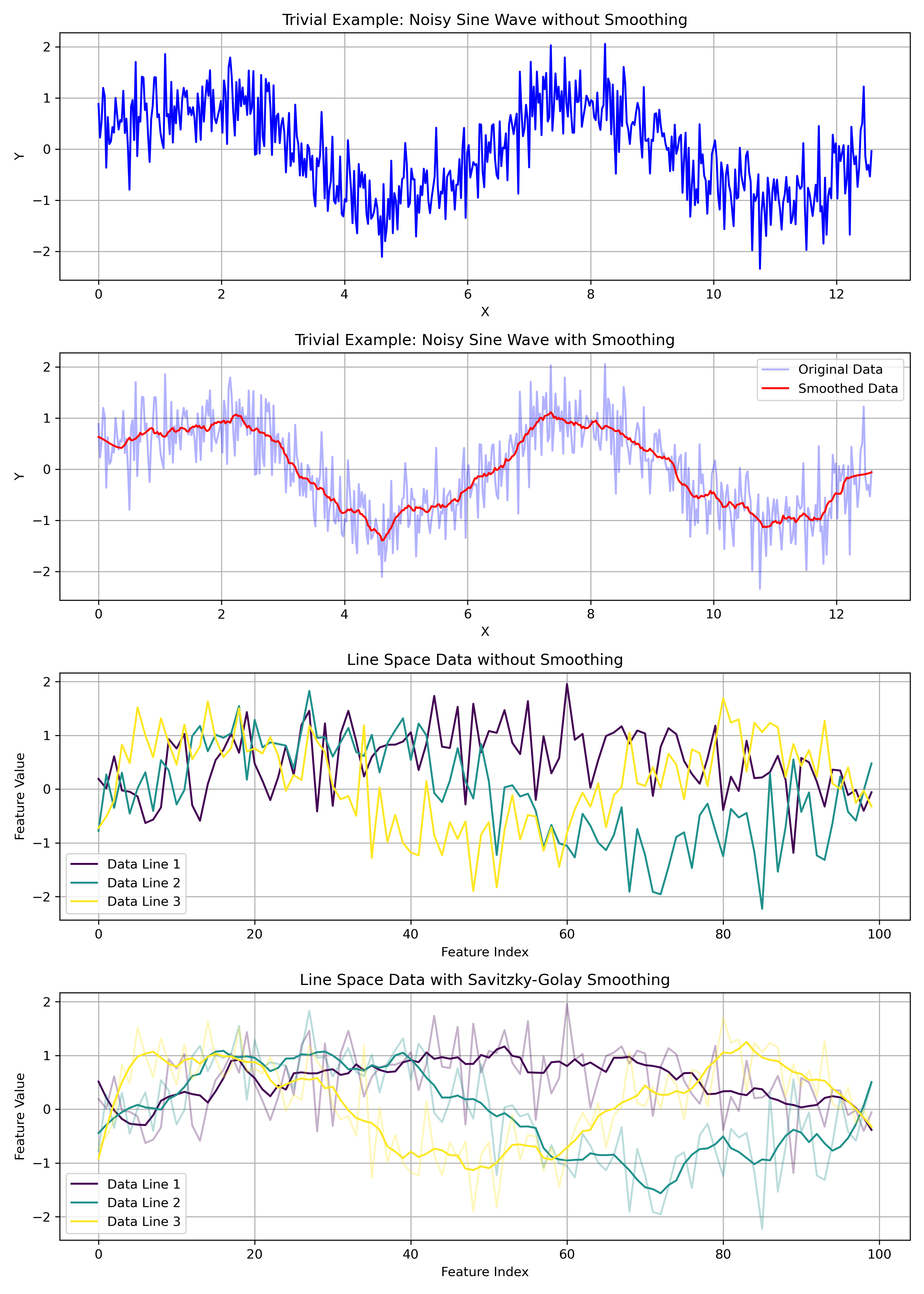}
\caption{Effect of Savitzky-Golay smoothing on data lines.}
\label{fig:smoothing_effect}
\end{figure}

\subsection{Combined Distance Metric}
To effectively capture both the magnitude differences and shape similarities between data points, we introduce a combined distance metric that balances Euclidean and DTW distances:

\begin{equation}
D(\mathbf{x}_i, \mathbf{x}_j) = \alpha \cdot D_{\text{DTW}}(\mathbf{L}_i, \mathbf{L}_j) + (1 - \alpha) \cdot D_{\text{EUC}}(\mathbf{x}_i, \mathbf{x}_j)
\end{equation}

where:
\begin{itemize}
    \item $D_{\text{DTW}}(\mathbf{L}_i, \mathbf{L}_j)$ is the DTW distance between the lines $\mathbf{L}_i$ and $\mathbf{L}_j$, capturing shape similarity.
    \item $D_{\text{EUC}}(\mathbf{x}_i, \mathbf{x}_j)$ is the Euclidean distance between the data points, capturing magnitude differences.
    \item $\alpha \in [0, 1]$ is the weighting parameter that balances the importance of the two distances.
\end{itemize}

\subsubsection{Interpretation of Alpha Parameter}
The parameter $\alpha$ controls the emphasis on shape similarity versus magnitude differences:

\begin{itemize}
    \item \textbf{Alpha Close to 1}: The distance metric focuses more on shape similarity, giving higher importance to the DTW distance. This is suitable when the pattern of feature values is crucial.
    \item \textbf{Alpha Close to 0}: The distance metric emphasizes magnitude differences, relying more on Euclidean distance. This is appropriate when absolute feature values are important.
\end{itemize}

Adjusting $\alpha$ allows the algorithm to adapt to different data characteristics and application needs. We recommend experimenting with values for best results.

\subsection{Algorithm Description}
The LSC algorithm consists of the following steps:

\begin{enumerate}
    \item \textbf{Data Standardization}: Standardize the dataset to have zero mean and unit variance to ensure all features contribute equally.
    \item \textbf{Smoothing (Optional)}: Apply the Savitzky-Golay filter to reduce noise and smooth each data line.
    \item \textbf{Initialization}: Randomly select $k$ data lines as initial cluster centers.
    \item \textbf{Distance Computation}: For each data line, compute the combined distance to each cluster center using Equation (10).
    \item \textbf{Cluster Assignment}: Assign each data line to the nearest cluster center based on the combined distance.
    \item \textbf{Cluster Center Update}: Update each cluster center by computing the median of the assigned data lines, enhancing robustness to outliers.
    \item \textbf{Convergence Check}: Repeat steps 4-6 until convergence (i.e., cluster centers stabilize) or a maximum number of iterations is reached.
\end{enumerate}

\subsubsection{Data Standardization}
Standardizing the data ensures that each feature contributes equally to the distance computations. It involves subtracting the mean and dividing by the standard deviation:

\begin{equation}
x_{ij}' = \frac{x_{ij} - \mu_j}{\sigma_j}
\end{equation}

where $\mu_j$ and $\sigma_j$ are the mean and standard deviation of feature $j$.

\subsubsection{Cluster Center Update Using Median}
Using the median instead of the mean when updating cluster centers may prove effective against outliers. The median minimizes the sum of absolute deviations, making it less sensitive to extreme values.

\begin{equation}
\mathbf{c}_j = \text{median} \{ \mathbf{x}_i \mid \mathbf{x}_i \text{ assigned to cluster } j \}
\end{equation}

\begin{figure}[h]
\centering
\includegraphics[width=3.6 in]{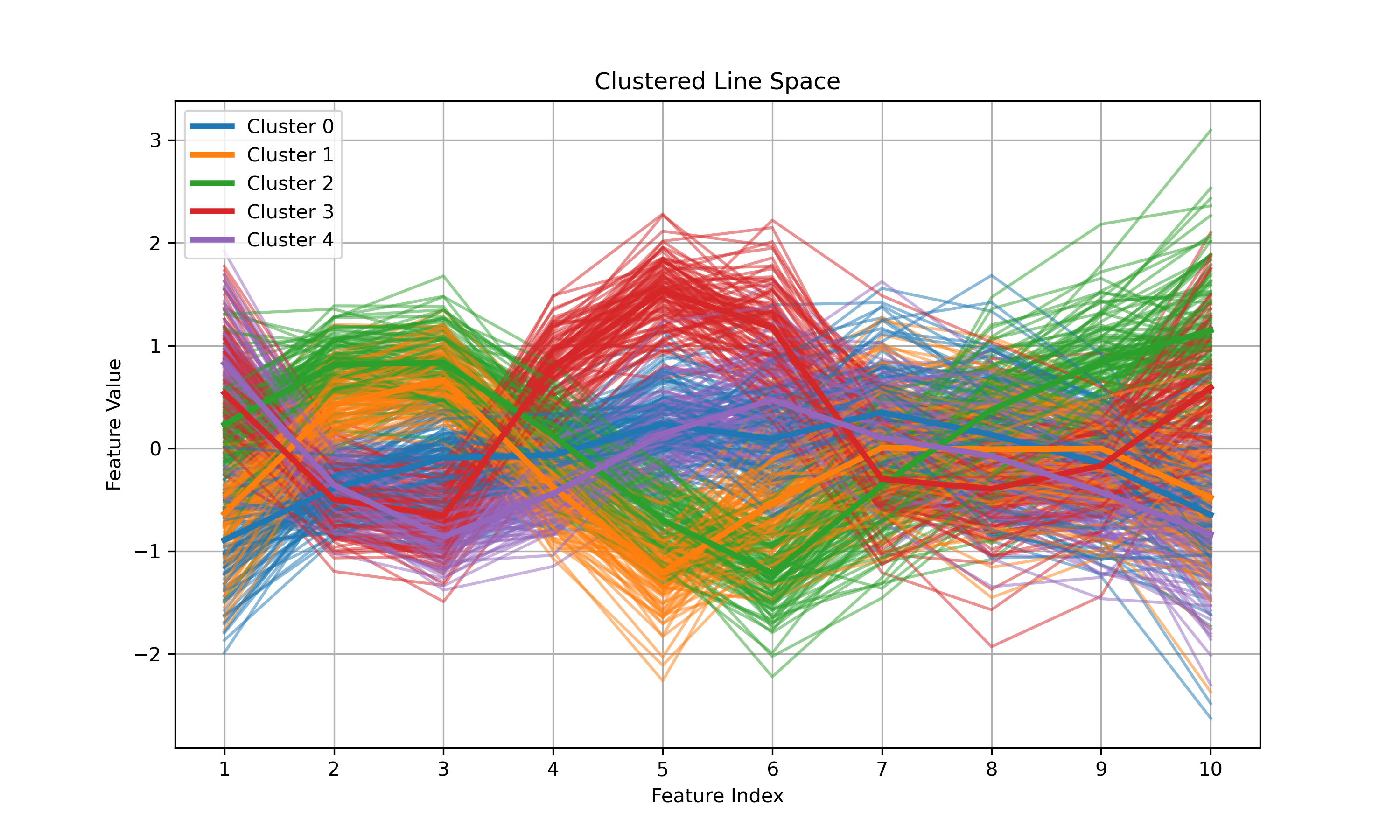}
\caption{Clustered line space after applying LSC. Different colors represent different clusters, k was chosen as 5.}
\label{fig:clustered_line_space}
\end{figure}

\section{Algorithm Implementation}
The LSC algorithm is formally presented in Algorithm \ref{alg:LSC}.

\begin{algorithm}[H]
\caption{Line Space Clustering (LSC)}\label{alg:LSC}
\begin{algorithmic}[1]
\REQUIRE Data matrix $\mathbf{X} \in \mathbb{R}^{n \times d}$, number of clusters $k$, weighting parameter $\alpha$, maximum iterations $max\_iter$, smoothing option $smoothing$
\ENSURE Cluster assignments $\mathbf{labels}$, cluster centers $\mathbf{C}$
\STATE \textbf{Standardize} the data $\mathbf{X}$
\IF{$smoothing$ is enabled}
    \STATE \textbf{Apply} Savitzky-Golay filter to smooth each data line
\ENDIF
\STATE \textbf{Initialize} cluster centers $\mathbf{C}$ by randomly selecting $k$ data lines
\FOR{$iter = 1$ to $max\_iter$}
    \FOR{each line point $\mathbf{x}_i$}
        \FOR{each cluster center $\mathbf{c}_j$}
            \STATE \textbf{Compute} $D(\mathbf{x}_i, \mathbf{c}_j)$ using Equation (10)
        \ENDFOR
        \STATE \textbf{Assign} $\mathbf{x}_i$ to cluster with nearest center
    \ENDFOR
    \STATE \textbf{Update} cluster centers $\mathbf{C}$ by computing median of assigned data lines
    \STATE \textbf{Check} for convergence
    \IF{converged}
        \STATE \textbf{Break}
    \ENDIF
\ENDFOR
\RETURN $\mathbf{labels}$, $\mathbf{C}$
\end{algorithmic}
\end{algorithm}

\section{Experimental Evaluation}

\subsection{Datasets}
We evaluated the performance of LSC on both synthetic and real-world datasets.

\subsubsection{Synthetic Datasets}
We generated synthetic datasets with varying numbers of samples, features, clusters, and noise levels to assess the algorithm's scalability and speed. Our synthetic datasets were also generated with known true class labels, thus allowing us to we employ external evaluation metrics. The synthetic data allows us to control the complexity and understand the behavior of LSC under different conditions. Furthermore, synthetic datasets were generated to resemble real datasets.

\paragraph{Dataset Generation}
The datasets were generated using Gaussian distributions with predefined means and covariances to form distinct clusters. Noise was introduced by adding Gaussian noise with varying standard deviations.

\paragraph{Parameters}
\begin{itemize}
    \item \textbf{Number of Samples}: 100000.
    \item \textbf{Number of Features}: 1024.
    \item \textbf{Number of Clusters}: 5.
    \item \textbf{Noise Levels}: 1, 2, 3, 5, 10.
\end{itemize}

\subsubsection{Real-World Datasets}
We utilized publicly available benchmark datasets sourced from the UCI Machine Learning Repository \cite{Dua2019} to evaluate the efficacy of the proposed methodology:

\begin{itemize}
\item \textbf{Iris Dataset}: This classical dataset comprises 150 instances, each characterized by four distinct morphological features, representing three species of Iris flowers (Setosa, Versicolor, and Virginica).
\item \textbf{Wine Dataset}: The dataset contains 178 samples of wines derived from three different cultivars, described by 13 chemical attributes pertinent to wine classification.
\end{itemize}

\subsection{Evaluation Metrics}
To rigorously assess the clustering performance, the following widely-adopted metrics were employed:

\subsubsection{Adjusted Rand Index (ARI)}
The Adjusted Rand Index (ARI) \cite{Hubert1985ComparingPartitions} quantifies the agreement between the clustering solution and the ground truth, adjusting for random chance. ARI values range between -1 and 1, where 1 denotes perfect clustering correspondence, and values near zero indicate clustering similarity equivalent to random labeling.

\subsubsection{Adjusted Mutual Information (AMI)}
Adjusted Mutual Information (AMI) \cite{Vinh2010InformationTheoreticMeasures} modifies the mutual information metric to correct for chance. It quantifies the degree of agreement between clustering outcomes and true classifications, with values approaching 1 representing ideal agreement and those nearing zero signifying no better than chance-level correspondence.

\subsubsection{Homogeneity Score}
Homogeneity \cite{Rosenberg2007Vmeasure} measures the extent to which clusters contain only elements from a single class. Scores range from 0 to 1, where 1 indicates perfectly homogeneous clusters exclusively composed of data points from the same class, and scores approaching 0 signify significant class mixing within clusters.

\subsubsection{Completeness Score}
Completeness \cite{Rosenberg2007Vmeasure} assesses the extent to which all instances belonging to the same class are grouped into the same cluster. A completeness score of 1 indicates that all members of each class are fully assigned to a single cluster, whereas lower scores imply fragmentation of classes across multiple clusters.

\subsubsection{V-Measure}
The V-Measure \cite{Rosenberg2007Vmeasure} is computed as the harmonic mean of homogeneity and completeness. It provides an integrated measure of clustering effectiveness, with a score of 1 indicating optimal clustering and scores approaching zero reflecting poor performance.

\subsubsection{Silhouette Coefficient}
The Silhouette Coefficient \cite{Rousseeuw1987Silhouettes} evaluates the cohesion and separation of clusters by measuring the similarity of each data point to its own cluster relative to other clusters. Values range from -1 to 1, where higher positive values indicate well-defined clusters with high cohesion and separation, values close to zero indicate overlapping clusters, and negative values suggest incorrect cluster assignments.

\subsection{Baseline Methods}
We compared LSC with traditional clustering algorithms:

\begin{itemize}
    \item \textbf{K-means}
    \item \textbf{Agglomerative Hierarchical Clustering}
    \item \textbf{DBSCAN}
    \item \textbf{Spectral Clustering}
\end{itemize}

\subsection{Implementation Details}
We implemented LSC in Python using NumPy and SciPy libraries. For DTW computation, we used the FastDTW algorithm to reduce computational complexity. All experiments were conducted on a machine with an Intel Core i7 processor and 16 GB RAM.

\subsection{Parameter Settings}
We set the parameters as follows unless specified:

\begin{itemize}
    \item \textbf{Number of Clusters ($k$)}: Set to the true number of clusters in the datasets.
    \item \textbf{Alpha ($\alpha$)}: We varied the value between 0.1 and 0.9 to observe the impact on clustering performance. Conclusively, the results depend on the data but for general purpose results we went with a value ranging from 0.25-0.75 (used in the table results).
    \item \textbf{Maximum Iterations}: 100. Assuming convergence would occur within 100 iterations.
    \item \textbf{Smoothing}: Enabled.
\end{itemize}

\subsection{Results}
In all the tables in bold are the highest (best) values for each metric for each clustering algorithm implemented.
\\
\subsubsection{Synthetic Data Results}

Table \ref{tab:synthetic_results} presents the clustering performance metrics on synthetic datasets with varying noise levels. 

\begin{table}[h]
\caption{Clustering Performance Metrics on Synthetic Datasets with Varying Noise Levels}
\centering
{\fontsize{8.1 pt}{9 pt}\selectfont  
\begin{tabular}{llccccc}
\toprule
\textbf{Noise} & \textbf{Metric} & \textbf{LSC} & \textbf{KM} & \textbf{Agglo} & \textbf{DBSCAN} & \textbf{Spectral} \\
\midrule
\multirow{6}{*}{1.0}
& ARI           & 0.9315         & \textbf{1.0000} & \textbf{1.0000} & 0.9240         & 0.6131 \\
& AMI           & 0.9347         & \textbf{1.0000} & \textbf{1.0000} & 0.9452         & 0.8170 \\
& Homogeneity   & 0.9162         & \textbf{1.0000} & \textbf{1.0000} & \textbf{1.0000} & 0.7792 \\
& Completeness  & 0.9563         & \textbf{1.0000} & \textbf{1.0000} & 0.9001         & 0.8662 \\
& V-measure     & 0.9358         & \textbf{1.0000} & \textbf{1.0000} & 0.9474         & 0.8204 \\
& Silhouette    & 0.6453         & \textbf{0.6716} & \textbf{0.6716} & 0.5711         & 0.1887 \\
\midrule
\multirow{6}{*}{2.0}
& ARI           & 0.8620         & 0.8923         & 0.9616         & N/A            & \textbf{0.9774} \\
& AMI           & 0.8538         & 0.9024         & 0.9618         & N/A            & \textbf{0.9771} \\
& Homogeneity   & 0.8427         & 0.9038         & 0.9624         & N/A            & \textbf{0.9775} \\
& Completeness  & 0.8705         & 0.9044         & 0.9625         & N/A            & \textbf{0.9776} \\
& V-measure     & 0.8564         & 0.9041         & 0.9624         & N/A            & \textbf{0.9775} \\
& Silhouette    & 0.3711         & \textbf{0.4214} & 0.4200         & N/A            & 0.4206 \\
\midrule
\multirow{6}{*}{3.0}
& ARI           & 0.7843         & 0.6104         & 0.6403         & N/A            & \textbf{0.8598} \\
& AMI           & 0.6965         & 0.7053         & 0.7538         & N/A            & \textbf{0.8763} \\
& Homogeneity   & 0.6925         & 0.7091         & 0.7569         & N/A            & \textbf{0.8779} \\
& Completeness  & 0.7113         & 0.7116         & 0.7592         & N/A            & \textbf{0.8789} \\
& V-measure     & 0.7018         & 0.7103         & 0.7581         & N/A            & \textbf{0.8784} \\
& Silhouette    & 0.2874         & \textbf{0.2876} & 0.2624         & N/A            & 0.2673 \\
\midrule
\multirow{6}{*}{5.0}
& ARI           & \textbf{0.3939} & 0.2954         & 0.2896         & N/A            & 0.2763 \\
& AMI           & \textbf{0.4309} & 0.3984         & 0.4035         & N/A            & 0.4158 \\
& Homogeneity   & \textbf{0.4391} & 0.4065         & 0.4084         & N/A            & 0.4188 \\
& Completeness  & \textbf{0.4423} & 0.4113         & 0.4196         & N/A            & 0.4335 \\
& V-measure     & \textbf{0.4407} & 0.4089         & 0.4139         & N/A            & 0.4260 \\
& Silhouette    & 0.2105         & 0.1914         & 0.1600         & N/A            & \textbf{0.2129} \\
\midrule
\multirow{6}{*}{10.0}
& ARI           & \textbf{0.1064} & 0.0761         & 0.0729         & N/A            & 0.0751 \\
& AMI           & \textbf{0.1592} & 0.1351         & 0.1140         & N/A            & 0.1260 \\
& Homogeneity   & \textbf{0.1637} & 0.1491         & 0.1274         & N/A            & 0.1398 \\
& Completeness  & \textbf{0.1745} & 0.1510         & 0.1318         & N/A            & 0.1426 \\
& V-measure     & \textbf{0.1641} & 0.1500         & 0.1295         & N/A            & 0.1412 \\
& Silhouette    &  \textbf{0.2038}         &  0.1823 & 0.1296         & N/A            & 0.1710 \\
\bottomrule
\end{tabular}
}
\label{tab:synthetic_results}
\end{table}

Notice, the N/A results for DBSCAN. The non-adjusted/adapted, basic DBSCAN did not find any clusters within the data.

\subsubsection{Real-World Data Results}
Table \ref{tab:realworld_results} shows the ARI and silhouette scores on real-world datasets.

\begin{table}[H]
\caption{Clustering Performance on Iris and Wine Datasets}
\centering
\begin{tabular}{llccccc}
\toprule
\multirow{2}{*}{\textbf{Dataset}} & \multirow{2}{*}{\textbf{Metric}} & \textbf{LSC} & \textbf{KM} & \textbf{Agglo} & \textbf{DBSCAN} & \textbf{Spectral} \\
 & & & & & & \\
\midrule
\multirow{6}{*}{Iris}
 & ARI & \textbf{0.6779} & 0.4328 & 0.6153 & 0.4421 & 0.6451 \\
 & AMI & \textbf{0.7045} & 0.5838 & 0.6713 & 0.5052 & 0.6856 \\
 & Silhouette & 0.4573 & \textbf{0.4799} & 0.4467 & 0.3565 & 0.4630 \\
 & V-measure & \textbf{0.7081} & 0.5896 & 0.6755 & 0.5114 & 0.6895 \\
 & Homogeneity & \textbf{0.7038} & 0.5347 & 0.6579 & 0.5005 & 0.6824 \\
 & Completeness & \textbf{0.7125} & 0.6570 & 0.6940 & 0.5228 & 0.6968 \\
\midrule
\multirow{6}{*}{Wine} 
 & ARI & 0.8961 & \textbf{0.8975} & 0.7899 & N/A & 0.4446 \\
 & AMI & \textbf{0.8757} & 0.8746 & 0.7842 & N/A & 0.5662 \\
 & Silhouette & 0.2818 & \textbf{0.2849} & 0.2774 & N/A & 0.2486 \\
 & V-measure & \textbf{0.8770} & 0.8759 & 0.7865 & N/A & 0.5725 \\
 & Homogeneity & \textbf{0.8795} & 0.8788 & 0.7904 & N/A & 0.4689 \\
 & Completeness & \textbf{0.8746} & 0.8730 & 0.7825 & N/A & 0.7348 \\
\bottomrule
\end{tabular}
\label{tab:realworld_results}
\end{table}

\subsubsection{Clustering Visualizations}
\begin{figure}[!ht]
    \centering
    \includegraphics[width=2.9 in]{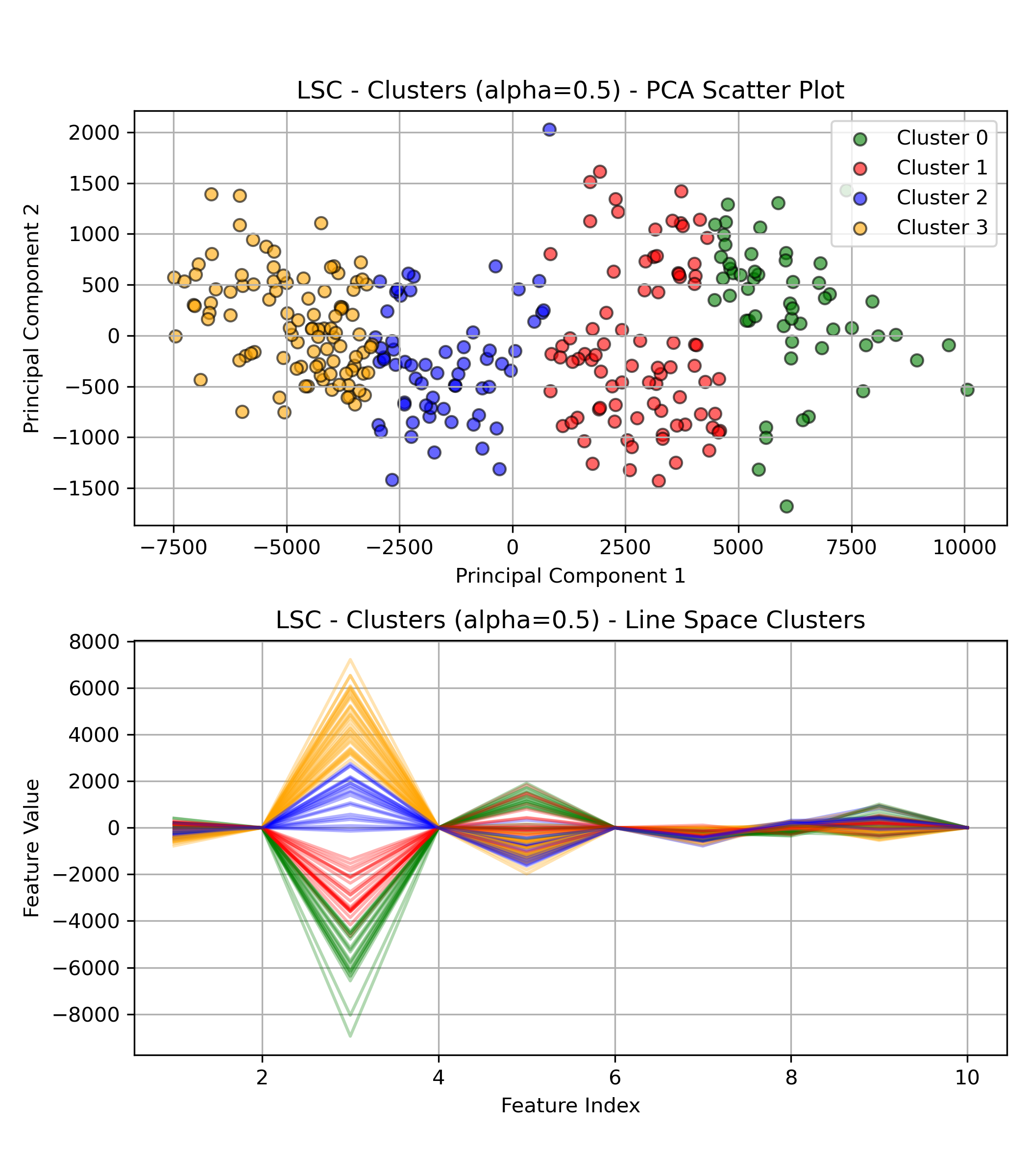}
    
    \includegraphics[width=2.55 in]{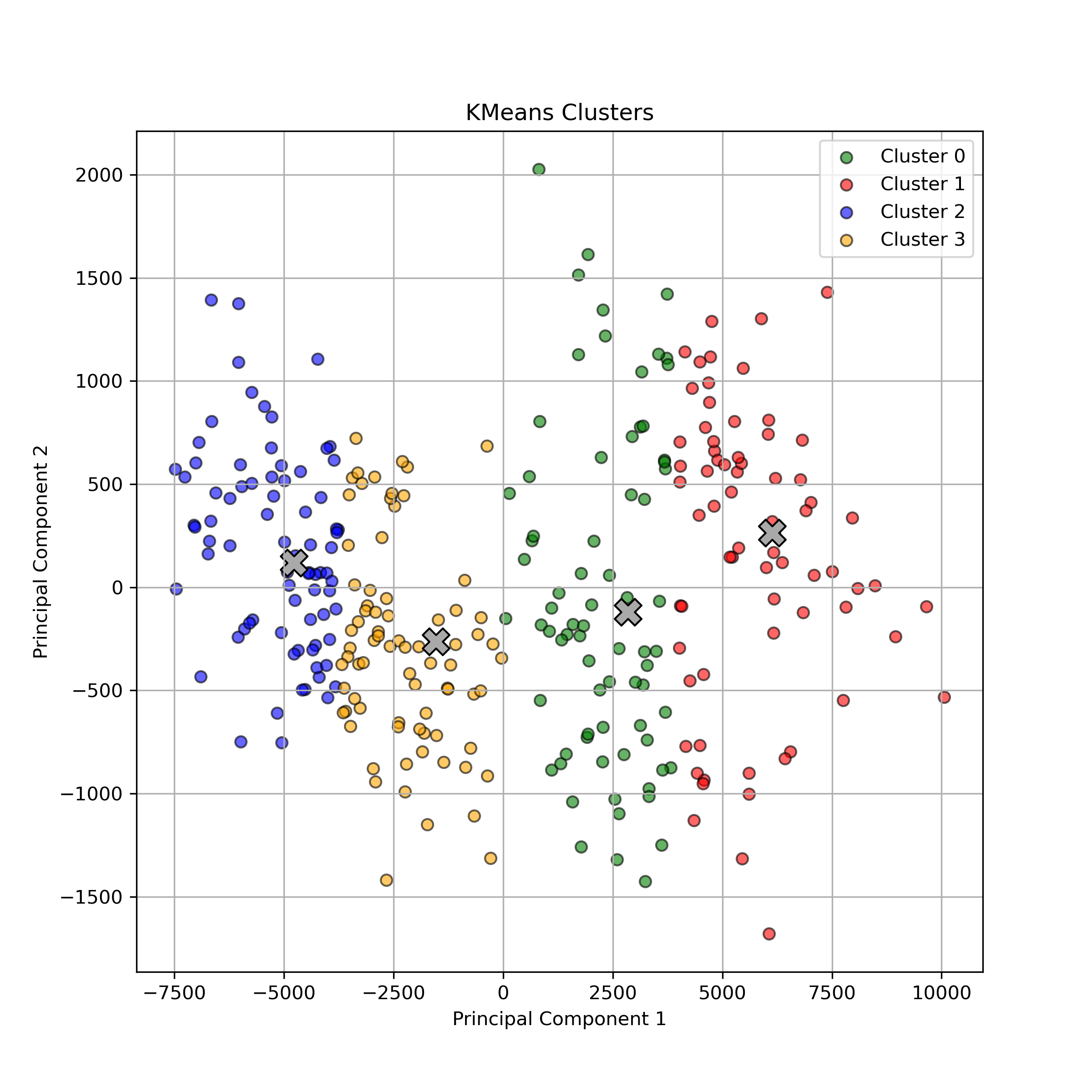}
    
    \includegraphics[width=2.55 in]{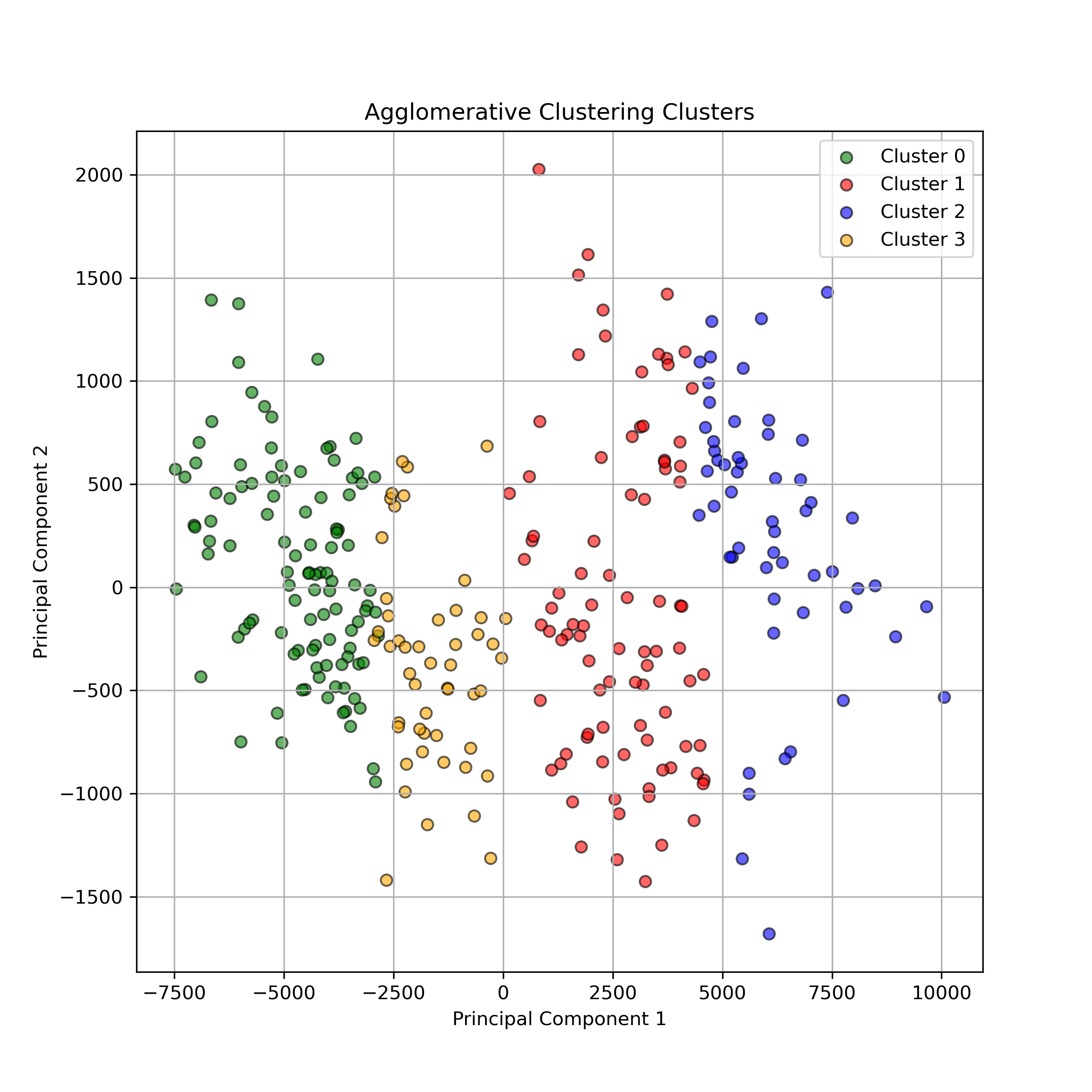}
    
    \caption{Sub-figure 1 shows LSC clusters 
    (in both 2D and Line Space), Sub-figure 2 shows K-means clusters, Sub-figure 3 shows Agglomerative clusters.}
    \label{fig:xd}
\end{figure}

To illustrate the clustering results, we provide visualizations comparing LSC with other algorithms in Figure \ref{fig:xd}. We can see similar results.

\subsubsection{Execution Time Comparison}
\begin{figure}[h]
\centering
\includegraphics[width=3 in]{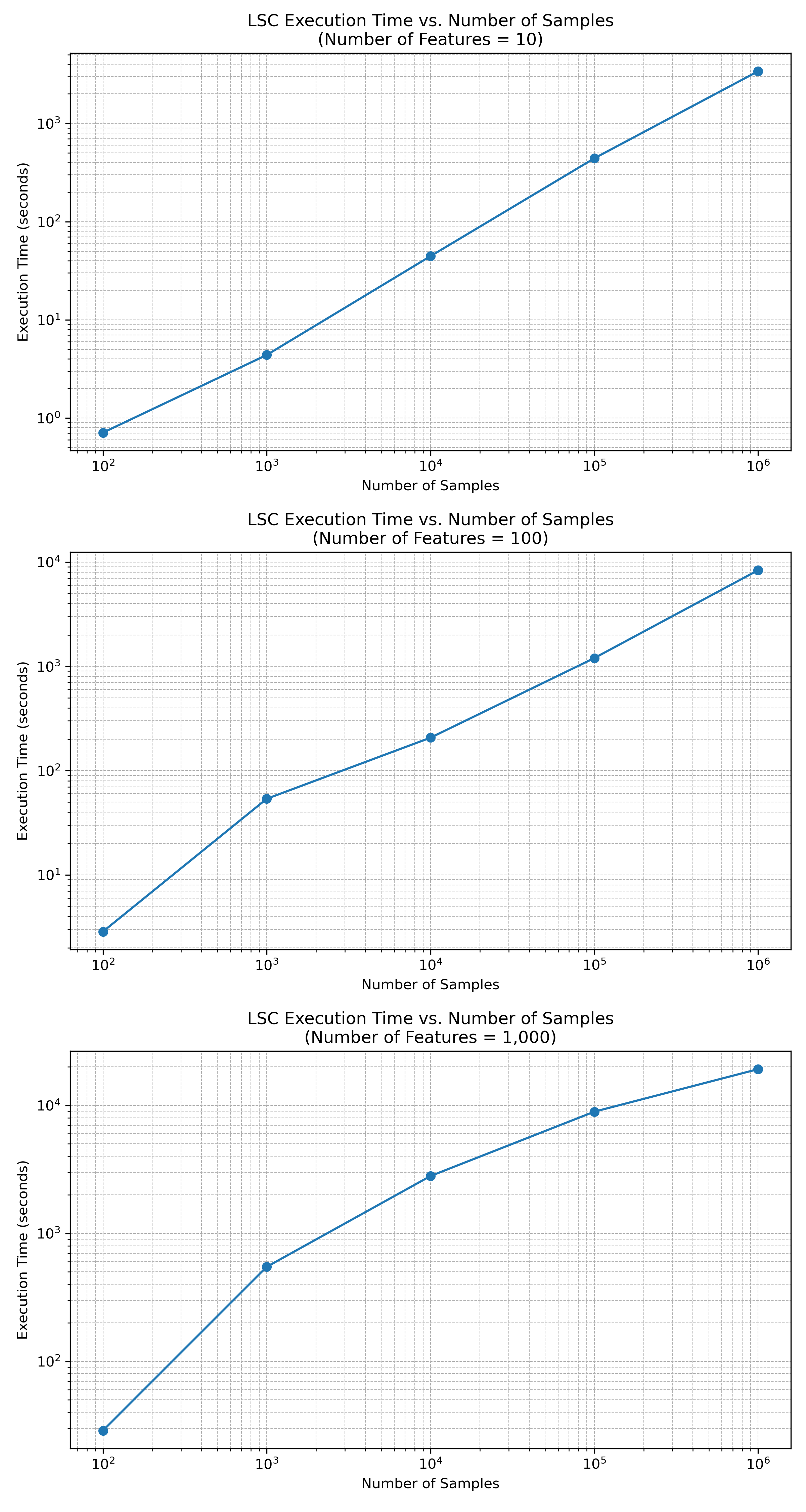}
\caption{Execution Time Comparison on arbitrarily created Datasets with a high noise index. LSC demonstrates reasonable execution times considering its complexity. Demonstrating an increase in execution timings in a linear fashion with data complexity.}
\label{fig:execution_time}
\end{figure}

Figure \ref{fig:execution_time} compares the execution times of LSC on varying feature numbers and samples.

\subsection{Effect of Smoothing}
We also investigated the impact of the smoothing step on clustering performance. Table \ref{tab:smoothing_effect} shows the ARI scores with and without smoothing a synthetic dataset with noise of 10.

\begin{table}[H]
\caption{Effect of Smoothing on Clustering Performance}
\centering
\begin{tabular}{lcc}
\toprule
\textbf{Method} & \textbf{With Smoothing} & \textbf{Without Smoothing} \\
\midrule
LSC (ARI)       & \textbf{ 0.85766}& 0.834644 \\
\bottomrule
\end{tabular}
\label{tab:smoothing_effect}
\end{table}

\section{Discussion}

\subsection{Analysis of Results}
The experimental results demonstrate that LSC consistently preforms as well as traditional clustering algorithms across various datasets and noise levels.

In the synthetic datasets, LSC maintained high ARI scores even as noise levels increased, whereas other algorithms showed performance degradation. With this we present LSC's ability to capture the underlying patterns in the data despite noise interference. As the noise level reached 10 only LSC and Spectral clustering remained competitive.

\subsection{Effectiveness of Line Space Transformation}
By transforming data points into lines, LSC leverages the sequential nature of feature values, enabling the detection of patterns and trends within individual data points. This approach allows for a more nuanced comparison between data points, beyond mere distance measurements.

The line space visualization (Figure \ref{fig:line_space}) illustrates how data points with similar patterns form clusters in the line space. After applying LSC, the clustered line space (Figure \ref{fig:clustered_line_space}) shows clear separation between clusters.

\subsection{Role of Dynamic Time Warping}
DTW plays a crucial role in measuring the similarity of the shapes of the data lines. By allowing non-linear alignments, DTW accounts for shifts and distortions in the feature value sequences, making LSC more robust to variability and noise.

However, DTW's computational complexity is a concern. While FastDTW reduces computation time, it may still be significant for very large datasets. The execution time comparison (Figure \ref{fig:execution_time}) shows that LSC is slower than algorithms like K-means but remains practical if computational and time resources are met.

\subsection{Impact of Alpha Parameter}
The weighting parameter $\alpha$ provides flexibility in balancing shape and magnitude similarities. Our experiments suggest that a medium $\alpha$ value (e.g., 0.2 to 0.5) offers the best performance for the datasets tested. Although we firmly recommend to adjust $\alpha$ based on the specific noise characteristics and to experiment with values.
Additionally, selecting the optimal $\alpha$ parameter may require domain knowledge or cross-validation, adding to the complexity of applying the algorithm.

\section{Conclusion}
We presented Line Space Clustering (LSC), an approach that effectively tries to clusters high-dimensional and noisy data by transforming data points into lines, via feature sequencing and using a combined distance metric that balances Euclidean and DTW distances. Our experiments demonstrate that LSC may outperform traditional clustering methods in some cases, particularly in noisy environments, or at random where unseen temporal feature patterns are present.
The detailed explanations of DTW and the Savitzky-Golay filter provide insights into how these components contribute to the algorithm's effectiveness.

\subsection{Future Work}
Future work includes optimizing the algorithm for large-scale datasets, possibly by incorporating parallel computing techniques or more efficient approximations of DTW. Developing methods for automatic selection of the $\alpha$ parameter could also enhance the algorithm's usability.

Exploring the application of LSC to other domains, such as bio-informatics or finance, may reveal further benefits of this approach.
Replicating our results on further datasets or very detailed comparisons would also prove useful.

\bibliographystyle{IEEEtran}
\bibliography{references}

\begin{thebibliography}{10}
\providecommand{\url}[1]{#1}
\csname url@samestyle\endcsname
\providecommand{\newblock}{\relax}
\providecommand{\bibinfo}[2]{#2}
\providecommand{\BIBentrySTDinterwordspacing}{\spaceskip=0pt\relax}
\providecommand{\BIBentryALTinterwordstretchfactor}{4}
\providecommand{\BIBentryALTinterwordspacing}{\spaceskip=\fontdimen2\font plus
\BIBentryALTinterwordstretchfactor\fontdimen3\font minus \fontdimen4\font\relax}
\providecommand{\BIBforeignlanguage}[2]{{%
\expandafter\ifx\csname l@#1\endcsname\relax
\typeout{** WARNING: IEEEtran.bst: No hyphenation pattern has been}%
\typeout{** loaded for the language `#1'. Using the pattern for}%
\typeout{** the default language instead.}%
\else
\language=\csname l@#1\endcsname
\fi
#2}}
\providecommand{\BIBdecl}{\relax}
\BIBdecl

\bibitem{Xu2015SurveyClustering}
D.~Xu and Y.~Tian, ``A comprehensive survey of clustering algorithms,'' \emph{Annals of Data Science}, vol.~2, no.~2, pp. 165--193, 2015.

\bibitem{Zhang2021ImageSegmentationReview}
Y.-J. Zhang, ``Image segmentation: A review of recent advances,'' \emph{Pattern Recognition}, vol. 120, p. 108102, 2021.

\bibitem{Steinbach2004HighDimensional}
M.~Steinbach, L.~Ertöz, and V.~Kumar, ``The challenges of clustering high dimensional data,'' pp. 273--309, 2004.

\bibitem{Zimek2012SurveyHighDimClustering}
A.~Zimek, E.~Schubert, and H.-P. Kriegel, ``A survey on unsupervised outlier detection in high-dimensional numerical data,'' \emph{Statistical Analysis and Data Mining}, vol.~5, no.~5, pp. 363--387, 2012.

\bibitem{Arthur2007KMeansPlusPlus}
D.~Arthur and S.~Vassilvitskii, ``k-means++: The advantages of careful seeding,'' in \emph{Proceedings of the Eighteenth Annual ACM-SIAM Symposium on Discrete Algorithms (SODA)}.\hskip 1em plus 0.5em minus 0.4em\relax Society for Industrial and Applied Mathematics, 2007, pp. 1027--1035.

\bibitem{Murtagh2012HierarchicalReview}
F.~Murtagh and P.~Contreras, ``Algorithms for hierarchical clustering: An overview,'' \emph{Wiley Interdisciplinary Reviews: Data Mining and Knowledge Discovery}, vol.~2, no.~1, pp. 86--97, 2012.

\bibitem{Parsons2004SubspaceReview}
L.~Parsons, E.~Haque, and H.~Liu, ``Subspace clustering for high dimensional data: A review,'' in \emph{ACM SIGKDD Explorations Newsletter}, vol.~6, no.~1, 2004, pp. 90--105.

\bibitem{Aggarwal1999ProjectedClustering}
C.~C. Aggarwal, J.~L. Wolf, P.~S. Yu, C.~Procopiuc, and J.~S. Park, ``Fast algorithms for projected clustering,'' in \emph{Proceedings of the ACM SIGMOD International Conference on Management of Data}, 1999, pp. 61--72.

\bibitem{Agrawal2005CLIQUE}
R.~Agrawal, J.~Gehrke, D.~Gunopulos, and P.~Raghavan, ``Automatic subspace clustering of high dimensional data for data mining applications,'' in \emph{Proceedings of the ACM SIGMOD International Conference on Management of Data}, 1998, pp. 94--105.

\bibitem{VonLuxburg2007SpectralClustering}
U.~Von~Luxburg, ``A tutorial on spectral clustering,'' \emph{Statistics and Computing}, vol.~17, no.~4, pp. 395--416, 2007.

\bibitem{Shi2000SpectralClustering}
J.~Shi and J.~Malik, ``Normalized cuts and image segmentation,'' \emph{IEEE Transactions on Pattern Analysis and Machine Intelligence}, vol.~22, no.~8, pp. 888--905, 2000.

\bibitem{Ester1996DBSCAN}
M.~Ester, H.-P. Kriegel, J.~Sander, and X.~Xu, ``A density-based algorithm for discovering clusters in large spatial databases with noise,'' in \emph{Proceedings of the Second International Conference on Knowledge Discovery and Data Mining (KDD)}, 1996, pp. 226--231.

\bibitem{Ankerst1999OPTICS}
M.~Ankerst, M.~M. Breunig, H.-P. Kriegel, and J.~Sander, ``Optics: Ordering points to identify the clustering structure,'' in \emph{ACM SIGMOD Record}, vol.~28, no.~2, 1999, pp. 49--60.

\bibitem{Aghabozorgi2015TimeSeriesClustering}
S.~Aghabozorgi, A.~S. Shirkhorshidi, and T.~Y. Wah, ``Time-series clustering–a decade review,'' \emph{Information Systems}, vol.~53, pp. 16--38, 2015.

\bibitem{Keogh2005DTWReview}
E.~Keogh and C.~A. Ratanamahatana, ``Exact indexing of dynamic time warping,'' \emph{Knowledge and Information Systems}, vol.~7, no.~3, pp. 358--386, 2005.

\bibitem{Berndt1994DTWOriginal}
D.~J. Berndt and J.~Clifford, ``Using dynamic time warping to find patterns in time series,'' \emph{KDD Workshop}, vol.~10, no.~16, pp. 359--370, 1994.

\bibitem{Salvador2007FastDTW}
S.~Salvador and P.~Chan, ``Toward accurate dynamic time warping in linear time and space,'' \emph{Intelligent Data Analysis}, vol.~11, no.~5, pp. 561--580, 2007.

\bibitem{Savitzky1964Smoothing}
A.~Savitzky and M.~J.~E. Golay, ``Smoothing and differentiation of data by simplified least squares procedures,'' \emph{Analytical Chemistry}, vol.~36, no.~8, pp. 1627--1639, 1964.

\bibitem{Schafer2011SavitzkyGolayReview}
R.~W. Sch{\"a}fer, ``What is a savitzky-golay filter?'' \emph{IEEE Signal Processing Magazine}, vol.~28, no.~4, pp. 111--117, 2011.

\bibitem{Dua2019}
\BIBentryALTinterwordspacing
D.~Dua and C.~Graff, ``{UCI Machine Learning Repository},'' 2019. [Online]. Available: \url{http://archive.ics.uci.edu/ml}
\BIBentrySTDinterwordspacing

\bibitem{Hubert1985ComparingPartitions}
L.~Hubert and P.~Arabie, ``Comparing partitions,'' \emph{Journal of Classification}, vol.~2, no.~1, pp. 193--218, 1985.

\bibitem{Vinh2010InformationTheoreticMeasures}
N.~X. Vinh, J.~Epps, and J.~Bailey, ``Information theoretic measures for clusterings comparison: Variants, properties, normalization and correction for chance,'' \emph{Journal of Machine Learning Research}, vol.~11, pp. 2837--2854, 2010.

\bibitem{Rosenberg2007Vmeasure}
A.~Rosenberg and J.~Hirschberg, ``V-measure: A conditional entropy-based external cluster evaluation measure,'' in \emph{Proceedings of the 2007 Joint Conference on EMNLP-CoNLL}, 2007, pp. 410--420.

\bibitem{Rousseeuw1987Silhouettes}
P.~J. Rousseeuw, ``Silhouettes: A graphical aid to the interpretation and validation of cluster analysis,'' \emph{Journal of Computational and Applied Mathematics}, vol.~20, pp. 53--65, 1987.

\end{thebibliography}

\end{document}